\documentclass[runningheads]{llncs}
\usepackage{graphicx}
\graphicspath{{./images/}}
\usepackage{comment}
\usepackage{amsmath,amssymb} 
\usepackage{color}

\usepackage[width=122mm,left=12mm,paperwidth=146mm,height=193mm,top=12mm,paperheight=217mm]{geometry}

\usepackage{epsfig}
\usepackage{subcaption}

\usepackage{booktabs}

\usepackage[pagebackref=true,breaklinks=true,letterpaper=true,colorlinks,bookmarks=false]{hyperref}

\usepackage{color}
\definecolor{redcol}{rgb}{1, 0, 0}
\definecolor{bluecol}{rgb}{0, 0, 1}

\usepackage[symbol]{footmisc}

\begin{document}
\pagestyle{headings}
\mainmatter
\def\ECCVSubNumber{1984}  

\title{Spatial Priming for Detecting Human-Object Interactions} 

\titlerunning{Spatial Priming for HOI Detection}
%
\author{Ankan Bansal* \and
    Sai Saketh Rambhatla* \and
Abhinav Shrivastava \and
\\Rama Chellappa}
\authorrunning{A. Bansal et al.}
%
\institute{University of Maryland, College Park, MD, USA \\
\email{\{ankan,rssaketh,abhinav,rama\}@umiacs.umd.edu}}
\maketitle

\footnotetext[1]{Denotes equal contribution.}

\begin{abstract}
   The relative spatial layout of a human and an object is an important cue for determining how they
   interact. However, until now, spatial layout has been used just as side-information for detecting
   human-object interactions (HOIs). In this paper, we present a method for exploiting this spatial
   layout information for detecting HOIs in images. The proposed method consists of a layout module
   which primes a visual module to predict the type of interaction between a human and an object.
   The visual and layout modules share information through lateral connections at several stages.
   The model uses predictions from the layout module as a prior to the visual module and the
   prediction from the visual module is given as the final output. It also incorporates semantic
   information about the object using word2vec vectors. The proposed model reaches an mAP of
   $24.79\%$ for HICO-Det dataset which is about $2.8\%$ absolute points higher than the current
   state-of-the-art. 
\end{abstract}

\section{Introduction}
\label{sec:intro}

\begin{figure}
    \centering
    \includegraphics[scale=0.27]{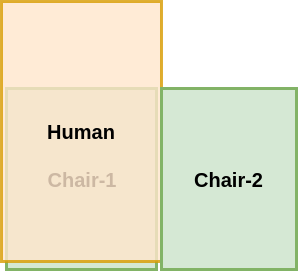}~
    \includegraphics[scale=0.20]{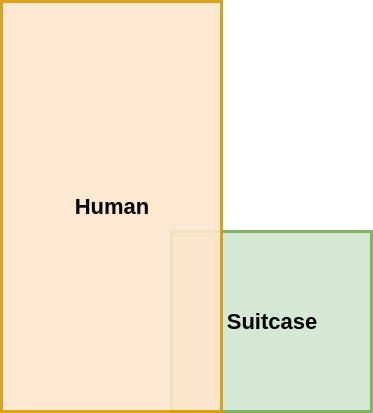}~
    \includegraphics[scale=0.21]{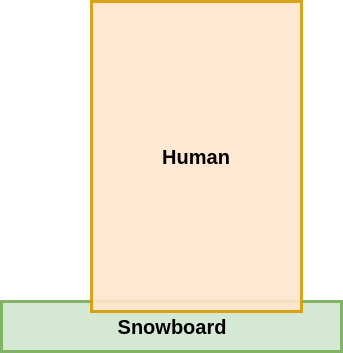}
    \caption{
    The relative spatial relationship between a human and an object provides much information about
    their interaction. We can infer that in the left image, the human is probably \texttt{sitting}
    on Chair-1 and \texttt{not interacting} with Chair-2. In the middle, the human might be
    \texttt{dragging} the suitcase. And the person on the right is probably \texttt{riding} a
    snowboard. 
}
    \label{fig:pull}
    \vspace{-10pt}
\end{figure}

Detecting human-object interactions (HOIs) involves localizing the interacting humans and objects
and correctly predicting the type of interaction (predicate) between them. Humans can guess the type
of interaction with just a quick glance at an image by considering the relative locations of the
human and the object. For example, in figure \ref{fig:pull}, the person on the left is very likely
to be \texttt{sitting} on Chair-1 and \texttt{not interacting} with Chair-2. Similarly, the person
in the middle is probably \texttt{dragging} the suitcase and the human on the right is
\texttt{standing on} the snowboard and possibly \texttt{riding} it. This ability to use spatial
relationships helps us in making guesses and eliminating improbable predictions. With additional
visual information, we can refine these priors to give better predictions. The
relative spatial layout of the human and the object involved in a HOI is greatly informative and
should be  utilized properly for predicting the interaction.

Existing work on HOI detection has not significantly leveraged this insight. Relative spatial
locations are usually not given sufficient attention. Current approaches either use a small
hand-created feature \cite{gupta2018no} or binary maps called interaction patterns (IPs)
\cite{chao2017learning}. Using hand-created features has the potential downside of not being able to
encode the fine-grained spatial relationships between objects. This limitation can be overcome by
using interaction patterns, which are binary maps representing the locations of the human and the
object in a HOI. Binary masks for the human and the object, as shown in figure \ref{fig:pull} can be
useful for predicting a prior on the interaction. This can be refined by using more visual
information from the image. 

We build on this idea for HOI detection. We address the question: how can we utilize the spatial
locations of the entities to improve HOI detection? Our proposed approach consists of a layout
branch and a visual branch.  The layout module outputs a prediction which is used as a prior by the
visual module. This prior prediction primes the visual branch which then outputs the final
predictions. Priming the visual module using predictions from the layout module enables our model to
fully utilize the spatial layout of the human and the object. We treat the relative geometry of
these entities as high-quality cues. 

Our layout and visual modules share information at multiple stages. Such information sharing between
different modalities \cite{feichtenhofer2016spatiotemporal} and at different levels of a network
\cite{lin2017feature,shrivastava2016beyond} has been shown to make the models learn better
representations. Lateral connections provide a way to share information between modules processing
different types of information. For example, \cite{feichtenhofer2016spatiotemporal} proposed lateral
connections between motion and appearance branches for video action recognition. Our layout module
receives information from the visual branch through lateral connections in the model. This sharing
of information enables the layout module to make stronger predictions about the predicate.  We put
the proposed approach in context of prior work in section \ref{sec:related}.

We evaluate our proposed approaches on the challenging HICO-Det dataset \cite{chao2017learning}. In
section \ref{sec:experiments}, we first present results for a simple baseline algorithm which uses a
good object detector and already achieves state-of-the-art results for HOI detection. Our proposed
model reaches a mean average precision (mAP) of $24.79\%$ on the HICO-Det dataset, which is about
$2.8$ absolute points higher than current state-of-the-art. We also conduct extensive analysis of
our proposed method to tease out the reasons for these improvements.

Finally, we discuss some avenues for future research in section
\ref{sec:conclusion}.

The most important contributions of this paper are two fold: (1) propose spatial priming as a way to
incorporate spatial layout of the human and object for HOI detection; (2) propose a model for HOI
detection based on spatial priming and information sharing between a layout and a visual module. In
addition, we conduct extensive analysis and evaluation of the proposed model to isolate sources of
performance improvement and report state-of-the-art results.

\section{Approach}
\label{sec:approach}

\begin{figure}[t]
    \centering
    \includegraphics[width=\linewidth]{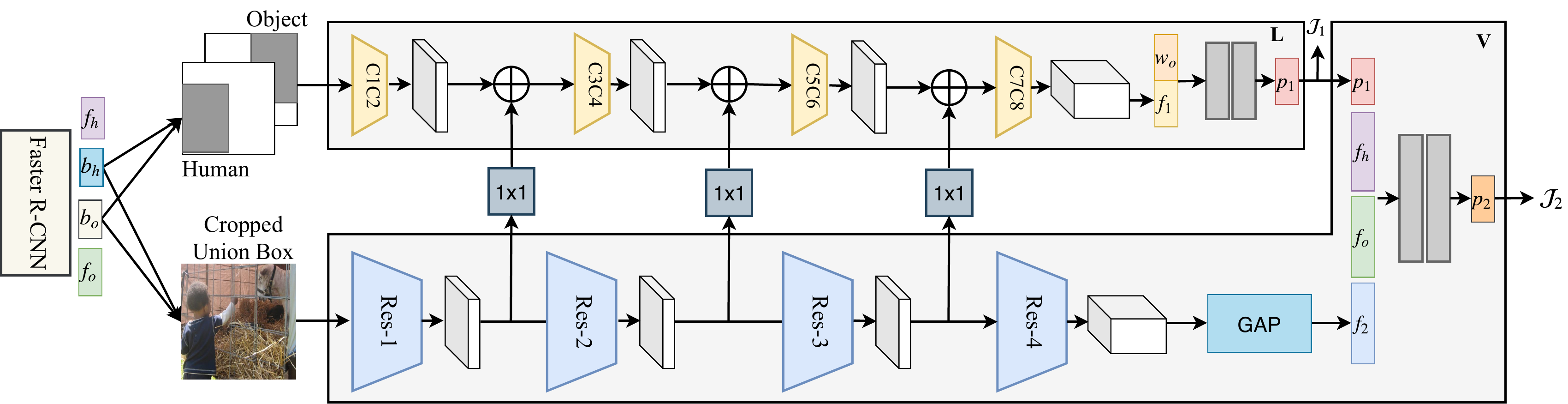}
    \vspace{-5pt}
    \caption{Proposed Pipeline. A Faster R-CNN object detector is used to detect humans and objects
    in an image. For each human-object pair, the interaction pattern, and union box are input to
\texttt{L} and \texttt{V} respectively. The predictions from \texttt{L} are used for priming
\texttt{V}. The visual module, \texttt{V} takes the union box, predictions from \texttt{L}, RoI
pooled human and object features from the object detector and outputs the final probabilities over
the predicates.}
    \label{fig:pipeline_ffs}
    \vspace{-10pt}
\end{figure}

The proposed model is composed of a relative layout module (\texttt{L}) and a visual module
(\texttt{V}) which share information at multiple stages. The predictions from \texttt{L} are used to
prime \texttt{V} and the final prediction is the output of \texttt{V}. Figure \ref{fig:pipeline_ffs}
shows the entire pipeline of our approach. Our model takes detections from an object detector in the
form of human-object pairs as input and outputs the probabilities for the predicates.  We describe
each component of our model next.

\subsection{Object Detector}
We start by using Faster R-CNN \cite{ren2015faster} to detect all humans and objects in an image and
create all candidate human-object pairs. Each pair has an associated human bounding box and an
object bounding box, giving a union box (the smallest bounding box covering both the human and the
object boxes). We crop the union box from the image and use it as an input to the visual module
\texttt{V}.

We generate binary spatial maps (interaction patterns) of the same dimensions as the union box for
the human and the object and stack these maps to produce a two-channel representation as shown in
figure \ref{fig:pipeline_ffs}. These spatial maps are used as inputs to the layout module
\texttt{L}. 

\subsection{Layout Module}
The spatial layout network, \texttt{L} is based on the idea that the relative layout and the
semantic category of the object can provide sufficient cues to determine the prior probabilities for
the interaction between a human and an object (see discussion in section \ref{sec:intro}).  We use a
shallow CNN as the layout network. This network takes the stacked spatial maps as input. We add
features from the visual module (described next) to intermediate layers in \texttt{L} via $1\times1$
convolutions. This provides visual context about the human and the object. After the final
convolution layer, we apply global average pooling to get the layout feature $f_1$. 

\noindent\textbf{Semantic knowledge.} Only the relative spatial layout might not be enough to
correctly determine the type of interaction. For example, in figure \ref{fig:pull}, it becomes
difficult to predict the relationship without the object identity. Therefore, we incorporate the
object identity in \texttt{L}. To include semantic information about the object, we concatenate
$f_1$ with the \texttt{word2vec} \cite{Mikolov2013DistributedRO} representation of the object, $w_o$
and pass the concatenated vector through two fully-connected layers which give the probabilities
over all predicates. Using semantic information about the object also helps in improving
generalization of the model to interactions involving previously unseen objects (zero-shot
detection). Using \texttt{word2vec} representations of the objects implicitly encodes semantic
similarities between objects. 

The output of the layout module are logits over the predicates which are used as an input to the
visual module. These logits act as a prior to the visual module which refines its output based on
this spatial prior.

\subsection{Visual Module}
The visual module \texttt{V} uses the predictions from \texttt{L} along with the visual information
from the cropped union bounding box to make the final prediction. We use a deeper network as the
base network in \texttt{V}. As mentioned above, intermediate features from \texttt{V} are added to
\texttt{L} to provide appearance and contextual information to the layout module. The base network
in \texttt{V} provides a feature vector $f_2$ after global average pooling the feature from the last
convolution layer.  We have two fully-connected layers at the end of the base convolution layers in
\texttt{V}. As input to these layers, we concatenate the features $f_2$, the prior predictions from
the layout module $p_1$, and the RoI-pooled human and object appearance features from the object
detector, $f_h$ and $f_o$. RoI-pooled features from the object detector provide explicit appearance
information about the human and the object. The output of the visual module is the final output of
the model.

\subsection{Lateral Connections}
We add features from intermediate layers in the visual module to intermediate layers in the layout
module via $1\times1$ convolution layers. Adding the visual features to features in the layout
module enables the model to explicitly share visual context not available in layout module.
Therefore, \texttt{L} can benefit from the appearance of the two interacting entities along with
their layout. This leads to a stronger spatial prior for the final stages of the visual module.
We will empirically demonstrate the importance of using this layout information in section
\ref{sec:experiments}.

\subsection{Spatial Priming}
Predictions from \texttt{L} based on the relative spatial layout prime the visual module. The layout
module \texttt{L} provides strong priors for the predicate which are refined by the visual module
which uses even more information about the human, object, and context appearance. Such priming
enables the visual module to gain from the information contained in the relative spatial layout of
the human and the object which is encoded by the binary spatial maps.

\subsection{Training}
We train \texttt{L} and \texttt{V} jointly. For both modules, we consider all predicates as
independent and use a weighted binary-cross entropy loss. The weights are simply inversely
proportional to the number of instances of the predicate in the dataset. The total loss is the sum
of the two losses from \texttt{L} and \texttt{V}.

Note that, we predict the probabilities/confidence scores of each
predicate for a human-object pair and not for the triplet \texttt{<human,predicate,object>}
directly. This gives our method the ability to detect previously unseen HOI categories (zero-shot
detection). To clarify, since we already
have the object labels from the object detector, we only need to output the predicate in order to
determine the interaction triplet. For example, the HICO-Det dataset contains $600$ annotated HOI
triplet categories of the form \texttt{<human,predicate,object>} but $117$ predicates. In total,
there are $9360$ ($117\times80$) possible interactions for this dataset. Only $600$ of these are
labeled. There might be more types of possible interactions than these. Our methods can potentially
detect such unlabeled HOI categories too (table \ref{tab:zero_shot}).

\section{Related Works}
\label{sec:related}

\textbf{Human-object interaction} (HOI) prediction being a special and important subset of visual
relationship prediction \cite{lu2016visual} is a well-studied problem. Early methods
\cite{gupta2007objects,yao2010modeling,yao2010grouplet,yao2011human,desai2012detecting} for HOI
prediction had mainly focused on developing hand-designed features and models. In particular, Yao
\emph{et al.} \cite{yao2010modeling} proposed a random field model which encodes the idea that humans poses
and objects can provide mutual context for each other. Delaitre \emph{et al.} \cite{delaitre2011learning}
built HOI features from spatial co-occurrences of body parts and objects. Hu \emph{et al.}
\cite{hu2013recognising} used exemplars in the form of density functions representing an HOI. All of
these are somewhat related to the proposed method owing to the use of the relative layout of humans
and objects to reason about HOIs.

More recently, Mallya \emph{et al.} \cite{mallya2016learning} used CNN features from local and
global context of a person along with a weighted loss to handle unbalanced training data. HO-RCNN
\cite{chao2017learning} and InteractNet \cite{gkioxari2017detecting} employed separate human,
    object, and interaction streams for HOI prediction. In particular, \cite{gkioxari2017detecting}
    jointly learned human and object detectors along with HOI detectors. However, these methods did
    not leave any scope for zero-shot HOI prediction. In \cite{xu2018interact}, Xu \emph{et al.}
    utilized gaze and pose information through a gaze-driven context-aware branch. Other methods
    \cite{gupta2018no,Zhou_2019_ICCV,Wan_2019_ICCV} also used human-pose  as fine-grained visual layout information. However,
    these methods require an additional model for predicting the pose. Unlike these, we argue that
    coarse relative layout along with the object identity provides sufficient cues to form a prior
    for interaction. We avoid the additional burden and potential errors of using a pose estimation
    model.

Several methods have utilized external \textbf{semantic knowledge for HOI prediction}
\cite{kato2018compositional,peyre2018detecting,bansal2019detecting}. In this work, we too have
used semantic information in the form of word vectors for object classes. These help in transferring
knowledge from an object to other similar object classes. Using semantic knowledge also helps in
generalizing to zero-shot HOI categories. Zero-shot HOI detection has previously been studied in
\cite{bansal2019detecting,shen2018scaling}. This followed several works on zero-shot object
recognition \cite{xian2017zero,kodirov2017semantic} and zero-shot object detection
\cite{bansal2018zero}.


Like the proposed model, Li \emph{et al.} \cite{li2018transferable} had also used priors for refining
predictions. However, they learned ``interactiveness priors" which only inform whether a human and
an object are interacting or not. We, instead, add a prior which informs about the HOI category
based on the relative spatial layout. 


\textbf{Relative spatial layout} is an important cue for predicting their interaction.
Some prior works have tried to incorporate the spatial relationship by encoding it as a small
hand-designed feature and passing it as input to a neural network
\cite{gupta2018no,bansal2019detecting}. Chao \emph{et al.}
\cite{chao2017learning} proposed ``interaction patterns" for encoding the relative spatial location.
Gao \emph{et al.} \cite{gao2018ican} also used such interaction patterns as a secondary branch of the model. However,
it can be argued
    that even these methods considered relative layout as secondary information to the visual
    features. In this paper, we use IPs as binary spatial maps to represent the relative layout of
    the human and the object. We present a principled approach for exploiting the information
    contained in such spatial maps. Our model uses a small CNN to combine visual, semantic, and
    geometric information to make a prediction for the predicate. This prediction is used as a prior
    for our final visual appearance-based branch.  

\textbf{Lateral connections} have been used for merging information from different
spatial resolutions \cite{lin2017feature}, for fusing optical and visual streams in two-stream
networks \cite{feichtenhofer2016spatiotemporal}, and for fusing coarse and
fine temporal resolutions \cite{feichtenhofer2018slowfast}. Shrivastava \emph{et al.}
\cite{shrivastava2016contextual} also used such lateral connections for priming an object detector
by contextual information from semantic segmentation. Feature Pyramid Network (FPN)
    \cite{lin2017feature} used lateral connections for building high-level semantic feature maps for
    object detection. Similarly, in \cite{shrivastava2016beyond} bottom-up
    and top-down pathways for object detection are connected using lateral connections.


%
%
%



\section{Experiments}
\label{sec:experiments}
We start with a brief description of the dataset and evaluation metrics and provide 
implementation details for our approach. We then discuss the model performance in fully-supervised
and zero-shot settings. Finally, we discuss and analyze the model through extensive ablation studies.

\subsection{Dataset and evaluation metrics} 
Following prior work, we use the challenging HICO-Det dataset
\cite{chao2017learning} for evaluating our approach. This dataset contains $600$ HOI triplet
\texttt{<human, predicate, object>} categories involving $117$ predicates and $80$ objects.  These
categories are divided into: (a) Rare - $138$ categories with less than $10$ training samples, and
(b) Non-rare categories. There are about $38,000$ training images containing about $120,000$
interactions and about $9,600$ test images with about $33,400$ HOIs.

Mean average precision (mAP) is used as the evaluation metric. A detected triplet is considered
correct if both human and object overlaps (IoU) with the ground truth are greater than $0.5$.
Performance is reported for the full set of 600 classes and also for the rare and non-rare classes
separately.  

Due to its inconsequential size ($<6,000$ training images and just
$26$ predicates), the V-COCO dataset \cite{gupta2015visual} does not provide any new insights into
HOI detection approaches.  Unsurprisingly, most recent state-of-the-art methods
\cite{bansal2019detecting,gupta2018no,peyre2018detecting} do not use V-COCO. To save space for
an exhaustive analysis of our model, we discuss V-COCO in the supplementary
material.

\subsection{Implementation Details} 
Following the state-of-the-art \cite{bansal2019detecting}, we start by fine-tuning a ResNet-101
\cite{He2016DeepRL} based Faster R-CNN \cite{ren2015faster} object detector for the HICO-Det dataset
\cite{chao2017learning}. The detector was originally trained on the COCO dataset
\cite{lin2014microsoft} which has the same $80$ object classes. Fine-tuning 
enables the detector to confidently detect objects more likely to be involved in an interaction. This
helps in improving the performance of downstream predicate classifiers. Please see supplementary
materials for details.

To create the training dataset, we consider all detections for which the detection confidence is
greater than $0.75$ and the overlap with a ground-truth human or object box is greater than $0.7$.
We create human-object pairs for each image using these detections and end up with about $250,000$
training HOI triplets.  For test proposals, we select only those object and human proposals which
have a confidence score greater than $0.9$ for a particular class.  This ensures that we get only
high confidence object detections and make fewer errors because of incorrectly detected objects and
humans. Each detection has an associated feature vector and bounding boxes. We use the human and
object bounding boxes, $b_h$ and $b_o$ respectively to compute the union box and the binary spatial
maps. 

For our model, the visual module is a ResNet-50 network and the layout module is a shallow 8-layer
CNN (see supplementary material for details). Each layer of \texttt{L} contains
a ReLU non-linearity and batch-normalization. We add lateral connections from each Residual block in
\texttt{V} to \texttt{L}, i.e., there are three lateral connections. Features from the residual
blocks, Res-1, Res-2, and Res-3 are added to the respective places in \texttt{L} as shown in figure
\ref{fig:pipeline_ffs}. The fully connected layers are of sizes $1024$ and $512$ in both \texttt{L}
and \texttt{V}. We reiterate that both \texttt{L} and \texttt{V} give the probabilities for the
$117$ predicates. This is unlike many previous methods which directly predict the HOI triplets
($600$ categories). We use the object labels from the object detector to output the final triplet.
This also enables us to detect previously unseen HOIs (zero-shot detection).

In all our experiments, we train the model for $10$ epochs with an initial learning rate of $0.1$
which is dropped by a tenth every $3$ epochs. Note that the object detector and the semantic
word-vectors are frozen while training our models, i.e., the detector needs to be trained only once.

\subsection{Results}

\begin{table}[t]
    \caption{Baseline results (mAP \%)}
    \vspace{-5pt}
    \begin{center}
    \resizebox{0.7\textwidth}{!}{
        \begin{tabular}{@{}l|c|c|c@{}}
            \toprule
             & \textbf{Full} & \textbf{Rare} & \textbf{Non-rare}\\
             \textbf{Method} & (600 classes) & (138 classes) & (462 classes) \\
            \midrule
            Baseline ResNet-50            & 20.80 & 15.63 & 22.34 \\
            Baseline ResNet-50$ + f_h + f_o$  & 21.49 & 14.43 & 23.60 \\  
            \bottomrule
        \end{tabular}
    }
    \end{center}
    \label{tab:results_baseline}
    \vspace{-20pt}
\end{table}

\noindent
\textbf{Strong Baseline.}
We start with a baseline CNN which predicts the predicates just based on the cropped union box. We
first use a ResNet-50 (R-50) network as the classifier which takes a cropped union box as input and
outputs the probabilities for each predicate. This network achieves an mAP of $20.80\%$ for the
HICO-Det test set. This is a strong albeit simple baseline which is already better than the current
state-of-the-art performance of $19.40\%$ (table \ref{tab:prior_work}). This reveals that the
existing methods can benefit from simplifying the algorithm and just using a better object detector
and a stronger feature extractor. A simple model like classifying the union box obtained from
detections from an object detector is enough to achieve better performance than existing methods.
Adopting the common practice \cite{li2018transferable,gkioxari2017detecting} of using the features
from the object detector, we append the RoI-pooled features to the features from the R-50, and
obtain an mAP of $21.49\%$. 
We summarize these results in table \ref{tab:results_baseline}.

\begin{table}
    \caption{Comparison with prior work. The performance (mAP \%) obtained by our method is
    significantly higher than existing methods.}
    \vspace{-5pt}
    \begin{center}
    \resizebox{0.65\textwidth}{!}{
        \begin{tabular}{@{}l|c|c|c@{}}
            \toprule
             & \textbf{Full} & \textbf{Rare} & \textbf{Non-rare}\\
             \textbf{Method} & (600 classes) & (138 classes) & (462 classes) \\
            \midrule
            Shen \emph{et al.} \cite{shen2018scaling} & 6.46 & 4.24 & 7.12 \\
            HO-RCNN + IP \cite{chao2017learning} & 7.30 & 4.68 & 8.08 \\
            HO-RCNN + IP + S \cite{chao2017learning} & 7.81 & 5.37 & 8.54 \\
            InteractNet \cite{gkioxari2017detecting} & 9.94 & 7.16 & 10.77 \\
            GPNN \cite{qi2018learning} & 13.11 & 9.34 & 14.23 \\
            iHOI \cite{xu2018interact} & 13.39 & 9.51 & 14.55 \\
            Xu \emph{et al.} \cite{Xu_2019_CVPR} & 14.70 & 13.26 & 15.13 \\
            ICAN \cite{gao2018ican} & 14.84 & 10.45 & 16.15 \\
            Wang \emph{et al.} \cite{Wang_2019_ICCV} & 16.24 & 11.16 & 17.75 \\
            Gupta \emph{et al.} \cite{gupta2018no} & 17.18 & 12.17 & 18.68 \\
            Interactiveness Prior \cite{li2018transferable} & 17.22 & 13.51 & 18.32 \\
            RPNN \cite{Zhou_2019_ICCV} & 17.35 & 12.78 & 18.71 \\
            PMFNet \cite{Wan_2019_ICCV} & 17.46 & 15.65 & 18.00 \\
            Peyre \emph{et al.} \cite{peyre2018detecting} & 19.40 & 15.40 & 20.75 \\
            Functional Gen. \cite{bansal2019detecting} & 21.96 & \textbf{16.43} & 23.62 \\
            \midrule
            Ours & \textbf{24.79} & 14.77 & \textbf{27.79} \\
            \bottomrule
        \end{tabular}
    }
    \end{center}
    \label{tab:prior_work}
    \vspace{-20pt}
\end{table}

\medskip
\noindent
\textbf{Comparison with Prior Work.} 
We compare the performance of our model with past work in table \ref{tab:prior_work}. Our model
achieves an mAP of $24.79\%$, which is over $2.8$ absolute percentage points higher than the current
state-of-the-art method \cite{bansal2019detecting} on the Full set of the HICO-Det dataset.  Our
method also performs about $4.2$ absolute percentage points better on Non-rare classes.
Interestingly, at the same time, even though we do not target them explicitly, our model achieves
competitive performance on Rare classes too. Note that the methods in \cite{bansal2019detecting} and
\cite{peyre2018detecting} are explicitly designed to target rare and unseen classes.

We also point out that, even using the original COCO detector instead of our fine-tuned detector,
our model achieves an mAP of $19.45\%$. This is the highest among all methods using an object
detector trained on COCO. In particular, the mAP achieved by the proposed method is significantly
higher ($2-12\%$ mAP) than previous methods
\cite{chao2017learning,gao2018ican,Wang_2019_ICCV} which aim to utilize the relative spatial layout
of the two entities. In addition, we obtain a higher performance than RPNN \cite{Zhou_2019_ICCV} and
PMFNet \cite{Wan_2019_ICCV} which use additional pose information using models trained on large
datasets. This demonstrates the strength of Spatial Priming as a way of modeling the
geometric layout.

\begin{table}
    \caption{Zero-shot HOI detection (mAP $\%$).}
    \vspace{-5pt}
    \begin{center}
    \resizebox{0.6\textwidth}{!}{
        \begin{tabular}{@{}l|c|c|c@{}}
            \toprule
             & \textbf{Unseen} & \textbf{Seen} & \textbf{All}\\
             \textbf{Method} & (120 classes) & (480 classes) & (600 classes)\\
            \midrule
            Shen \emph{et al.} \cite{shen2018scaling} & 5.62 & - & 6.26 \\
            Functional Gen. \cite{bansal2019detecting} & 10.93 & 12.60 & 12.26 \\
            \midrule
            Ours & \textbf{11.06} & \textbf{21.41} & \textbf{19.34} \\ 
            \bottomrule
        \end{tabular}
    }
    \end{center}
    \label{tab:zero_shot}
    \vspace{-20pt}
\end{table}

\medskip
\noindent
\textbf{Zero-Shot HOI Detection.} 
The proposed approach can help improve the performance for zero-shot HOI
detection. Table \ref{tab:zero_shot} compares the performance of our method with the
state-of-the-art methods \cite{shen2018scaling,bansal2019detecting} on zero-shot HOI detection.
Prior work divides the classes into a set of $120$ unseen and $480$ seen classes. We use
the same setting here. The model is trained with training data for only the seen classes and is
evaluated on the set of unseen classes. Note that the classes are divided such that the there is at
least one interaction involving each of the $80$ objects in the training set, i.e., the model is
trained with at least one HOI involving each object. From table \ref{tab:zero_shot} we observe that
our model achieves a higher mAP than \cite{bansal2019detecting} for Unseen classes while also
improving the mAP for Seen classes by a huge margin. We have used the same train-test splits as
\cite{bansal2019detecting}.

\subsection{Ablation Analysis}
We now extensively analyze our model in tables \ref{tab:combined_ablation},
\ref{tab:NP_ablation}, and \ref{tab:results_NL}. 

\begin{table}[t]
    \caption{Ablation studies for the model. We report mAP (\%) in each case. In all sub-tables ``Standard" refers to the model shown in
figure \ref{fig:pipeline_ffs}.}
    \vspace{-10pt}
    \begin{center}
            \begin{subtable}[t]{0.3\textwidth}
                \caption{\textbf{Effect of} $\mathbf{w_o}$. Row 2 is the standard model without
                $w_o$. Note that the performance without $w_o$ is lower than the
            Standard case. This is particularly true for the Rare classes.}
                \vspace{-5pt}
                \resizebox{\textwidth}{!}{
                \begin{tabular}{@{}l|c|c|c@{}}
                    \toprule
                    \textbf{Setting} & \textbf{Full} & \textbf{Rare} & \textbf{Non-rare}\\
                    \midrule
                    Standard  
                    & \textbf{24.79} & \textbf{14.77} & 27.79 \\ 
                    Standard - $w_o$ 
                    & 24.47 & 12.16 & \textbf{28.14} \\ 
                    \bottomrule
                \end{tabular}
                }
                \label{tab:ab1}
            \end{subtable}
            ~~~~
            \begin{subtable}[t]{0.3\textwidth}
                \caption{\textbf{Lateral connection methods}. Concat is the model with lateral
                additions replaced by concatenation. 3x3add uses $3\times3$ convs in lateral
            connections instead of $1\times1$ used in the Standard setting.}
                \vspace{-5pt}
                \resizebox{\textwidth}{!}{
                \begin{tabular}{@{}l|c|c|c@{}}
                    \toprule
                    \textbf{Setting} & \textbf{Full} & \textbf{Rare} & \textbf{Non-rare}\\
                    \midrule
                    Standard 
                    & \textbf{24.79} & \textbf{14.77} & \textbf{27.79} \\ 
                    Concat 
                       & 24.00 & 13.91 & 27.02 \\ 
                    3$\times$3add  
                       & 24.21 & 13.34 & 27.47 \\ 
                    \bottomrule
                \end{tabular}
                }
                \label{tab:ab2}
            \end{subtable}
            ~~~~
            \begin{subtable}[t]{0.3\textwidth}
                \caption{\textbf{Importance of \texttt{L}}. ImgCNNA-ImgR50 is the model where input
                to \texttt{L} is the cropped union box. Similarly, in ImgR50-ImgR50, the layout
            module is a ResNet-50 with the union box as input. (Standard is IPCNNA-ImgR50)}
                \vspace{-5pt}
                \resizebox{\textwidth}{!}{
                \begin{tabular}{@{}l|c|c|c@{}}
                    \toprule
                    \textbf{Setting} & \textbf{Full} & \textbf{Rare} & \textbf{Non-rare}\\
                    \midrule
                    Standard 
                    & \textbf{24.79} & \textbf{14.77} & \textbf{27.79} \\ 
                    ImgCNNA-ImgR50 
                       & 22.28 & 10.87 & 25.69 \\
                    ImgR50-ImgR50 
                           & 24.07 & 11.96 & 27.68 \\
                    \bottomrule

                \end{tabular}
                }
                \label{tab:ab3}
            \end{subtable} 

            \vspace{-10pt}

            \begin{subtable}[t]{0.4\textwidth}
                \caption{\textbf{Utility of } $\mathbf{f_h}$, $\mathbf{f_o}$. Concat contains
                    concatenated lateral connections.
                Standard$-f_h$$-f_o$ has no $f_h$ and $f_o$. Standard-Larger contains
            larger hidden layers and no $f_h$ and $f_o$. Similarly for Concat$-f_h$$-f_o$ and
        Concat-Larger.}
                \vspace{-5pt}
                \resizebox{0.9\textwidth}{!}{
                \begin{tabular}{@{}l|c|c|c@{}}
                    \toprule
                    \textbf{Setting} & \textbf{Full} & \textbf{Rare} & \textbf{Non-rare}\\
                    \midrule
                    Standard 
                    & \textbf{24.79} & \textbf{14.77} & 27.79 \\ 
                    Standard$-f_h$$-f_o$ 
                       & 22.32 & 13.14 & 25.07 \\ 
                    Standard-Larger 
                    & 24.60 & 13.58 & \textbf{27.89} \\ 
                    Concat 
                       & 24.00 & 13.91 & 27.02 \\
                    Concat$-f_h$$-f_o$ 
                       & 21.87 & 13.05 & 24.51 \\ 
                    Concat-Larger 
                       & 23.41 & 14.44 & 26.09 \\ 
                    \bottomrule
                \end{tabular}
                }
                \label{tab:ab4}
            \end{subtable}
            ~~~~~~~~~~
            \begin{subtable}[t]{0.4\textwidth}
                \centering
                \caption{\textbf{Different lateral connections}. Conn1 is the model with just one
                lateral connection from \texttt{V} to \texttt{L} which is at Res-1. Conn2 has just
            one lateral connection at Res-2 and Conn3 has the lateral connection at Res-3.
        \texttt{L}-\texttt{V} has all connections from \texttt{L} to \texttt{V}.}
                \vspace{-5pt}
                \resizebox{0.8\textwidth}{!}{
                \begin{tabular}{@{}l|c|c|c@{}}
                    \toprule
                    \textbf{Setting} & \textbf{Full} & \textbf{Rare} & \textbf{Non-rare}\\
                    \midrule
                    Standard 
                    & \textbf{24.79} & \textbf{14.77} & \textbf{27.79} \\ 
                    \texttt{L}-\texttt{V} 
                       & 23.81 & 13.44 & 26.91 \\ 
                    Conn1
                       & 23.63 & 10.75 & 27.48\\
                    Conn2
                       &  24.01&  12.86& 27.34\\ 
                    Conn3
                       &  22.87&  11.42& 26.28\\ 
                    \bottomrule
                \end{tabular}
                }
                \label{tab:ab5}
            \end{subtable}
    \end{center}
    \label{tab:combined_ablation}
    \vspace{-20pt}
\end{table}

\smallskip
\noindent
\underline{Importance of $w_o$.}
Word-vectors $w_o$ encode the semantic similarities between objects. Table \ref{tab:ab1} shows that
using the word-vector in the layout module leads to performance improvement. The complete model
which uses the \texttt{word2vec} vectors $w_o$ achieves an mAP $24.79\%$. Removing this word-vector
leads to a lower performance ($24.47\%$). 

\smallskip
\noindent
\underline{Type of Lateral Connection.}
Table \ref{tab:ab2} illustrates that adding the features from the visual module to the geometric
module achieves higher performance than concatenating the features. The mAP in the case of
addition of features is $24.79\%$.
Compare this to the mAP of $24\%$ when the features are concatenated instead. The reason for
this is that adding features from the visual module forces \texttt{L} to explicitly
focus on the human and object. This ensures that the relevant regions of the image
are given more importance. Similarly, when using $3\times3$ convolutions in the lateral connections
instead of $1\times1$, the performance is slightly lower. This is because using $3\times3$
convolutions increase the receptive field of the features. This dilutes the focus on the human and
the object which in turn leads to a lower performance. 

\smallskip
\noindent
\underline{Layout Module.}
The importance of using the relative spatial layout of the human and object is demonstrated using
the data in table \ref{tab:ab3}. The first row in the table is the standard case when the human and
object spatial maps are given as input to the shallow layout network. This model gives an mAP of
$24.79\%$. Now, if we remove the binary spatial maps (IP) and input the cropped union box image to
the layout module too, the performance of the model drops to just $22.28\%$ (second row). Note that
this model has the same number of parameters as the previous model. The only difference is the input
to \texttt{L}. To ensure that the drop in performance is not due to a weak layout network, we
replace the small CNN in the layout module with a ResNet-50 network. Again, the input to both the
layout and visual branches is the cropped union box. Even this model, with a much larger number of
parameters than the standard case, gives an mAP of just $24.07\%$. This shows that relative spatial
layout of the human and the object provides irreplaceable information for determining the type of
interaction.

\smallskip
\noindent
\underline{Importance of $f_h$ and $f_o$.}
From table \ref{tab:ab4}, we observe that a model gives a lower performance if appearance features
from the object detector are not used. For example, the Standard model reaches an mAP of $24.79\%$
while the Standard model trained without $f_h$ and $f_o$ reaches only $22.32\%$. Similarly, the
performance for the Concat model (from table \ref{tab:ab2}) goes down from $24\%$ to just $21.87\%$
on removing the features. Clearly, $f_h$, and $f_o$ help in achieving higher performance. Recall
that we had observed the same effect with the Baseline model (table \ref{tab:results_baseline}). 

To analyze if these improvements are because of a larger number of parameters, we removed $f_h$ and
$f_o$ and increase the sizes of the fully connected layers such that the number of trainable
parameters in this model and the Standard model are roughly the same. We call this model
Standard-Larger. This model gives an mAP of $24.60\%$. This is down from $24.79\%$ obtained by the
Standard model. Similarly, the Concat-Larger model gives an mAP of $23.41\%$, down from the Concat
model which gave $24\%$. So, even though some of the performance gain when using the appearance
features could be due to a larger number of parameters, it does not explain the whole difference. We
believe that the features $f_h$ and $f_o$ do, in fact, provide useful information for predicting the
HOI.

\smallskip
\noindent
\underline{Different Connections.}
We show that having lateral connections at multiple depths in the network is important for obtaining
a good performance. We study whether having just one lateral connection can be enough. From table
\ref{tab:ab5}, we infer that the answer is \textit{no}. Just one lateral connection after either 
Res-1 block, Res-2 block, and Res-3 block (rows 3, 4, and 5 respectively) gives worse performance
than having connections at all three places. In particular, having just one connection after Res-3
gives the lowest performance. This is because by this depth the visual module loses most spatial
information and the layout module does not benefit from adding visual features. This shows that
frequent information sharing between the two modules via lateral connections gives significant
performance improvements. We also observe that passing information from the spatial layout module to
the visual module also achieves a lower mAP. 

\begin{figure}
        \centering
        \includegraphics[width=0.9\linewidth]{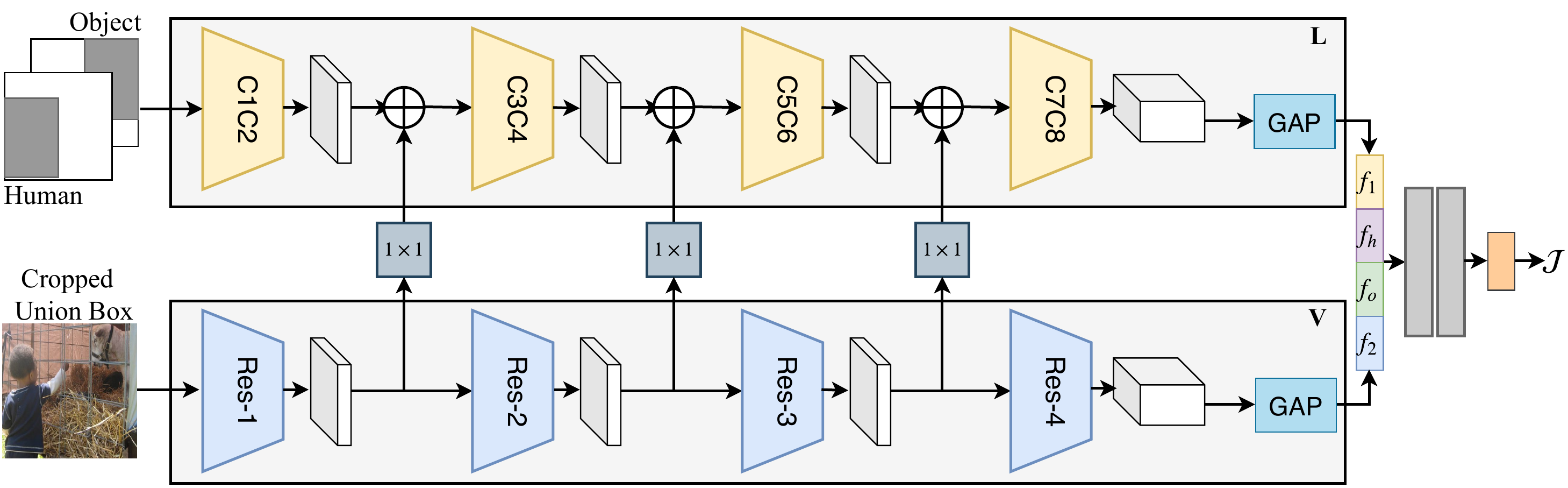}
        \vspace{-5pt}
        \caption{\textbf{No priming (NP)}. This model removes the spatial priming from our model
        (figure \ref{fig:pipeline_ffs}). Human and object bounding boxes from an object detector
    give the interaction patterns and the union box. Global Average Pooled (GAP) features from the
geometry and visual networks are concatenated to the human and object RoI pooled features from the
object detector. Two FC layers are used to get the probabilities/confidences over the predicates.}
        \label{fig:pipeline_NP}
        \vspace{-10pt}
\end{figure}

\medskip
To analyze the effect of each of the component in our model in more detail, we conduct ablation
studies in two further settings. First, we study the utility and behaviour of lateral connections
without priming. We remove the loss $\mathcal{J}_1$ and instead of adding layout priors, $p_1$ to
the visual module, we directly add the global average pooled features, $f_1$. This gives the model
shown in figure \ref{fig:pipeline_NP}. We call this model NP.

\medskip
\noindent
\textbf{No Priming.}
The first row in table \ref{tab:NP_ablation} gives the performance of the NP model (no priming)
shown in figure \ref{fig:pipeline_NP}. This model achieves an mAP of $23.41\%$ on the Full HICO-Det
dataset. Notice that this is higher than the Baseline model discussed earlier ($21.49\%$ table
\ref{tab:results_baseline}). This highlights the importance of the spatial layout even in this
simpler setting. 

\begin{table}
    \caption{\textbf{NP Results} (mAP \%). NP is the model shown in figure \ref{fig:pipeline_NP}
    with lateral connections from \texttt{V} to \texttt{L}. NC is the same model without lateral
connections. Similarly, \texttt{L}-\texttt{V} has connections from the layout branch to the visual
branch. \texttt{V}-\texttt{L}-concat concatenates the features from \texttt{V} and \texttt{L}
instead of adding.}
    \vspace{-5pt}
    \begin{center}
    \resizebox{0.55\textwidth}{!}{
        \begin{tabular}{@{}l|c|c|c@{}}
            \toprule
             & \textbf{Full} & \textbf{Rare} & \textbf{Non-rare}\\
             \textbf{Method} & (600 classes) & (138 classes) & (462 classes) \\
            \midrule
            \texttt{V}-\texttt{L}-add (NP)       & \textbf{23.41} & 12.14 & \textbf{26.78} \\ 
            NC        & 22.56 & \textbf{12.78} & 25.48 \\ 
            \texttt{L}-\texttt{V}     & 22.45 & 12.23 & 25.50 \\ 
            \texttt{V}-\texttt{L}-concat       & 22.76 & 11.78 & 26.04 \\ 
            \bottomrule
        \end{tabular}
    }
    \end{center}
    \label{tab:NP_ablation}
    \vspace{-20pt}
\end{table}

We further analyze the behavior of this model in different conditions. In table
\ref{tab:NP_ablation} \texttt{V}-\texttt{L} is the model with lateral connections from the visual
module (\texttt{V}) to the layout module (\texttt{L}). To illustrate the positive impact of these
lateral connections, we remove all lateral connections and train the resulting model. This model is
called ``NC" (no connection) in table \ref{tab:NP_ablation}. NC reaches an mAP of only $22.56\%$.
Clearly, lateral connections enable better utilization of the relative spatial layout of a person
and an object. However, note that this is still higher than the Baseline model ($21.49\%$), clearly
demonstrating that leveraging layout information is important for improving HOI detection
performance.

Further, we observe that connections from layout module to the visual module (\texttt{L}-\texttt{V}) give almost the same performance as having no connection (NC). 
Also, the final row in table \ref{tab:NP_ablation} is the case where we concatenate the intermediate features from the
visual module to the features of the layout module instead of adding. This model, though better than
having no connections, is still worse than the \texttt{V}-\texttt{L} model. We believe that in the
case of concatenation, the network does not learn to attend to the human and the object. On the
other hand, when we add the features instead, we explicitly force the network to attend to the human
and object regions of the image. This enables it to learn better mappings from human and object
appearances to the correct predicate. Recall that we had seen similar behavior in table
\ref{tab:ab2}.

\begin{figure}
        \centering
        \includegraphics[width=0.65\linewidth]{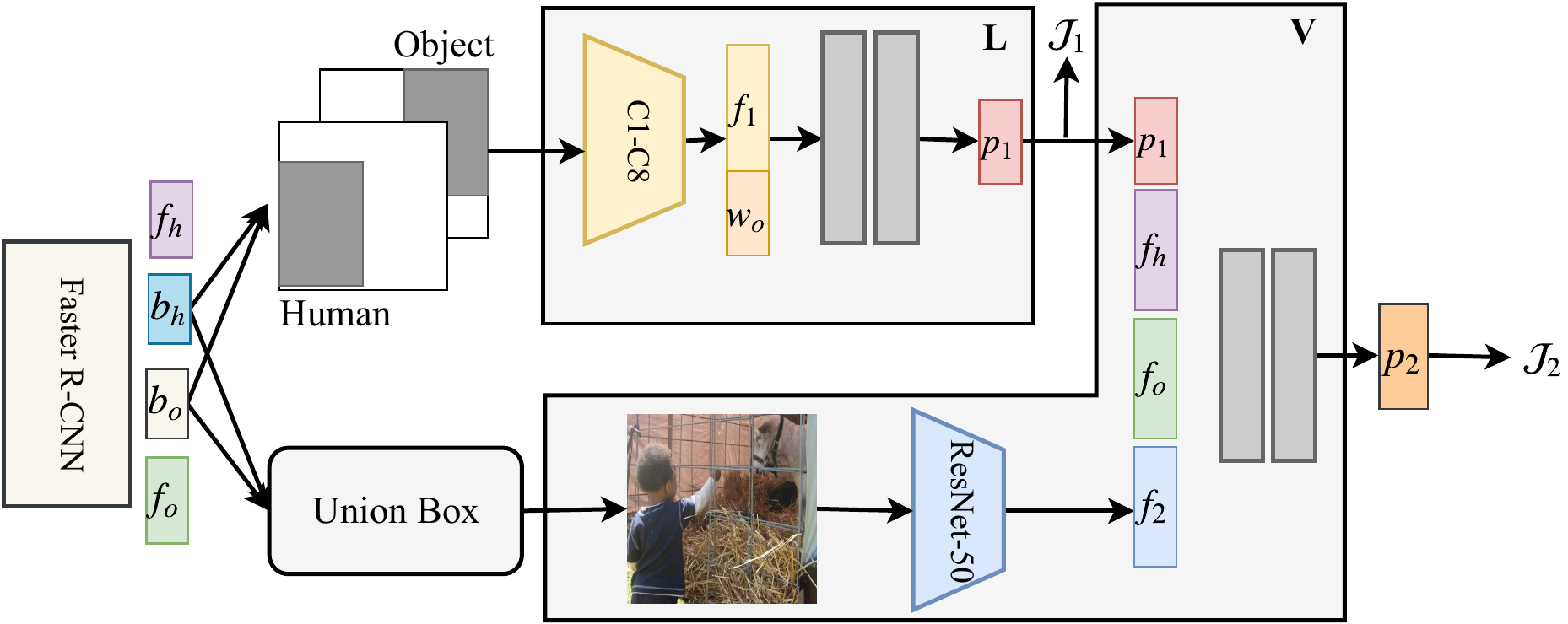}
        \vspace{-5pt}
        \caption{\textbf{No lateral connections (NL)}. We remove lateral connections from our model
        (figure \ref{fig:pipeline_ffs}). Now, \texttt{L} predicts the interaction just based on the
    spatial layout. These predictions are given as a prior to \texttt{V} which also uses the union
bounding box and the RoI pooled features from the object detector to make the final prediction.}
        \label{fig:pipeline_s1s2}
        \vspace{-20pt}
\end{figure}

\medskip
Next, we study the effect of removing lateral connections from our model. This is an important case
and will show how informative the layout module is on its own. This will help us pinpoint how the
components of our model behave in the presence of only spatial priming without lateral connections.
We remove the lateral connections from our model in figure \ref{fig:pipeline_ffs}. This gives us the
model shown in figure \ref{fig:pipeline_s1s2}. We call this model NL.

\medskip
\noindent
\textbf{No Lateral Connections.}
Results and ablation studies for the NL model are listed in table \ref{tab:results_NL}. The best
model reaches $23.90\%$ in mAP on the HICO-Det dataset. It contains a shallow layout branch
\texttt{L} which predicts the predicate based only on the spatial layout of the human and the
object. This prediction is used as a prior by the visual network \texttt{V} which gives the final
prediction. We highlight the performance of the layout network \texttt{L}. It achieves an mAP of
$18.35\%$ on the Full set of HICO-Det. This shows that there is significant information about the
interaction category contained in the relative spatial layout of the human and object. When properly
trained, using only this information might be better than most existing methods (13/15 methods in
table \ref{tab:prior_work}).

Again, the importance of a layout-based prior is apparent when comparing the performance of this
model with the performance of the baseline R-50 model which had reached only $21.49\%$ (table
\ref{tab:results_baseline}). The last row in table \ref{tab:results_NL} shows that removing the word
vector $w_o$ from the model leads to a drop in performance. This is driven down by the reduction in
the performance of the layout model \texttt{L} which went from $18.35\%$  in the usual case to just
$16.33\%$. Removing the appearance features from the detector, $f_h$ and $f_o$, also results in
lower performance. We had seen the same trends even in the presence of lateral connections in table
\ref{tab:combined_ablation}. Also, note that the performance for NL ($23.90\%$) is higher than the
performance for NP ($23.41\%$ table \ref{tab:NP_ablation}), showing that the idea of spatial priming
is a significant source of improvement achieved by our proposed model.

\vspace{-15pt}
\begin{table}
    \caption{\textbf{NL Results} (mAP \%). First row (NL) is the model shown in figure
    \ref{fig:pipeline_s1s2}. NL - $f_h$ - $f_o$ represents the model trained without the appearance
features from the object detector. NL - $w_o$ is NL without the word vector for the object.}
    \vspace{-10pt}
    \begin{center}
    \resizebox{0.6\textwidth}{!}{
        \begin{tabular}{@{}lc|c|c|c@{}}
            \toprule
            &  & \textbf{Full} & \textbf{Rare} & \textbf{Non-rare}\\
            \textbf{Method} & Model & (600 classes) & (138 classes) & (462 classes) \\
            \midrule
            NL & 
            \scriptsize{\texttt{L}}  & \scriptsize{18.35} & \scriptsize{8.20} & \scriptsize{21.38} \\ 
                             & \texttt{V}        & \textbf{23.90} & 10.82 & \textbf{27.81} \\ 
            \midrule
            NL - $f_h$ - $f_o$ & 
            \scriptsize{\texttt{L}}  & \scriptsize{17.44} & \scriptsize{10.14} & \scriptsize{19.62} \\ 
                             & \texttt{V}        & 23.19 & \textbf{14.71} & 25.72 \\ 
            \midrule
            NL - $w_o$ & 
            \scriptsize{\texttt{L}}  & \scriptsize{16.33} & \scriptsize{8.45} & \scriptsize{18.69} \\ 
                             & \texttt{V}        & 22.91 & 11.29 & 26.39 \\ 
            \bottomrule
        \end{tabular}
    }
    \end{center}
    \label{tab:results_NL}
    \vspace{-30pt}
\end{table}

\section{Discussion and Conclusion}
\label{sec:conclusion}

We discuss some avenues for further improvements and finally conclude.


\subsection{Discussion}
In this paper, we have not explicitly considered ways of improving detection for rare classes. The
competitive performance for rare classes in table \ref{tab:prior_work} is a by-product of our
approach, particularly, using semantic knowledge in the form of \texttt{word2vec} representations.
HOI datasets will always suffer from the long-tail problem. Future research should focus on
improving performance for rare classes. Clever class-weighing strategies and using more semantic
knowledge as in \cite{peyre2018detecting} could be some ways of going forward.

Another limitation of our method is the dependence on a pre-trained object detector. Future work
should also focus on jointly training the HOI prediction model and the object detector. Since HOI
detection and object detection have complementary objectives (a better object detector leads to
        better HOI detection), this line of approach could significantly improve performance for
both HOI detection and object detection.

\subsection{Conclusion}
We have presented an approach for using the relative layout information of a human and an object for
detecting interactions between them. Our proposed model consists of two modules: one for processing
the relative spatial layout of a human and an object, and the other for processing visual
information. The visual module is primed using the prediction of the layout module. We have
systematically analyzed the model and our experiments shown that this method can significantly
out-perform state-of-the-art methods for HOI detection.


\section*{Acknowledgement}
This project was supported by the Intelligence Advanced Research Projects Activity (IARPA) via
Department of Interior/Interior Business Center (DOI/IBC) contract number D17PC00345 and by  DARPA
via ARO contract number\\ W911NF2020009. The U.S.  Government is authorized to reproduce and
distribute reprints for Governmental purposes not withstanding any copyright annotation thereon.\\
\textbf{Disclaimer}: The views and conclusions contained
herein are those of the authors and should not be interpreted as necessarily representing the
official policies or endorsements, either expressed or implied of IARPA, DOI/IBC, DARPA, ARO, or the
U.S.  Government.

\clearpage
\setcounter{section}{0}
\renewcommand{\thesection}{S\arabic{section}}
\setcounter{table}{0}
\renewcommand{\thetable}{S\arabic{table}}
\setcounter{figure}{0}
\renewcommand{\thefigure}{S\arabic{figure}}

\section{Architecture of Layout Module}
The layout module is a shallow 8-layer CNN. Each layer of \texttt{L} contains a ReLU non-linearity
and batch-normalization. See details of the convolution layers in table \ref{tab:s1_architecture}.
We add lateral connections from each Residual block in \texttt{V} to \texttt{L}, i.e., there are
three lateral connections. Features from the residual blocks, Res-1, Res-2, and Res-3 are added to
the respective places in \texttt{L}.

\begin{table}
    \caption{\textbf{Architecture of \texttt{L}}. C1-C8 are convolution layers. Layer dimensions are
        in the shape \texttt{kernel\_width} $\times$ \texttt{kernel\_height} $\times$
        \texttt{output\_channels}. Numbers in parenthesis are
strides.}
    \vspace{-5pt}
    \begin{center}
        \resizebox{0.7\textwidth}{!}{
            \begin{tabular}{@{}c|c|c@{}}
                \toprule
                \textbf{Layer} & \textbf{Layer Dimensions} & \textbf{Output Sizes} \\
                \midrule
                C1C2 & 7$\times$7$\times$64 (2), MaxPool (2), 3$\times$3$\times$256 (1) & 56$\times$56$\times$256 \\
                C3C4 & 1$\times$1$\times$128 (1), 3$\times$3$\times$512 (2) & 28$\times$28$\times$512 \\
                C5C6 & 1$\times$1$\times$256 (1), 3$\times$3$\times$1024 (2) & 14$\times$14$\times$1024 \\
                C7C8 & 1$\times$1$\times$512 (1), 3$\times$3$\times$2048 (2) & 7$\times$7$\times$2048 \\
                GAP  & 7$\times$7 & 1$\times$1$\times$2048 \\
                FC1  & 1024 & 1024 \\
                FC2  & 512 & 512 \\
                \bottomrule
            \end{tabular}
        }
    \end{center}
    \label{tab:s1_architecture}
    \vspace{-20pt}
\end{table}

\section{V-COCO}
Similar to prior methods \cite{gkioxari2017detecting,Zhou_2019_ICCV,Wan_2019_ICCV} we use the 24
action classes involving a person and an object. We use our ResNet-101 object detector to extract
the human and object bounding boxes from each image. For generating the training set, we use all
proposals which overlap with a ground-truth entity box with an IoU greater than $0.5$. For testing,
we use human and object proposals with confidence $>0.8$. A major difference between the HICO-Det
dataset \cite{chao2017learning} and V-COCO is the absence of annotations for the
\texttt{no-interaction} or \texttt{background} class. We generate samples for
\texttt{no-interaction} by considering un-labeled human-object interactions as belonging to this
class.  Following the standard practice in object detection \cite{ren2015faster}, we use the
background and labeled classes in a ratio of 3:1.

Table \ref{tab:vcoco_prior} shows that the performance achieved by Spatial Priming (49.2\% mAP) is
significantly higher than most existing methods. In table \ref{tab:vcoco_class_wise}, we list the
class-wise AP obtained by our method for the 24 classes under consideration.


\begin{table}
    \caption{Comparison with prior work for the V-COCO dataset. The performance (mAP$_\text{role}$ \%) obtained by our method is
    higher than existing methods.}
    \vspace{-5pt}
    \begin{center}
    \resizebox{0.45\textwidth}{!}{
        \begin{tabular}{@{}l|c@{}}
            \toprule
            \textbf{Method} & \textbf{mAP$_\text{role}$} \\
            \midrule
            Gupta et al. \cite{gupta2015visual} & 31.8 \\
            InteractNet \cite{gkioxari2017detecting} & 40.0 \\
            GPNN \cite{qi2018learning} & 44.0 \\
            ICAN \cite{gao2018ican} & 45.3 \\
            RPNN \cite{Zhou_2019_ICCV} & 47.5 \\
            Wan et al. \cite{Wan_2019_ICCV} & 48.6 \\
            RP$_\text{T2}$C$_\text{D}$ \cite{li2018transferable} & 48.7 \\
            \midrule
            Spatial Priming (Ours) & \textbf{49.2} \\
            \bottomrule
        \end{tabular}
    }
    \end{center}
    \label{tab:vcoco_prior}
    \vspace{-20pt}
\end{table}

\begin{table}[h!]
    \caption{Class-wise AP obtained by our Spatial Priming approach for the V-COCO dataset.}
    \vspace{-5pt}
    \begin{center}
    \resizebox{0.4\textwidth}{!}{
        \begin{tabular}{@{}l|c@{}}
            \toprule
            \textbf{Class} & \textbf{AP$_\text{role}$}\\
            \midrule
            hold-obj & 36.81\\
            sit-instr & 32.94\\
            ride-instr & 64.34\\
            look-obj & 42.34\\
            hit-instr & 69.02\\
            hit-obj & 39.39\\
            eat-obj & 46.70\\
            eat-instr & 13.31\\
            jump-instr & 52.69\\
            lay-instr & 31.66\\
            talk\_on\_phone-instr & 30.15\\
            carry-obj & 37.45\\
            throw-obj & 41.72\\
            catch-obj & 52.94\\
            cut-instr & 34.04\\
            cut-obj & 48.30\\
            work\_on\_computer-instr & 64.85\\
            ski-instr & 46.03\\
            surf-instr & 76.43\\
            skateboard-instr & 86.68\\
            drink-instr & 45.01\\
            kick-obj & 78.37\\
            read-obj & 33.18\\
            snowboard-inst & 75.36\\
            \midrule
            Average Role AP & 49.15\\
            \bottomrule
        \end{tabular}
    }
    \end{center}
    \label{tab:vcoco_class_wise}
    \vspace{-20pt}
\end{table}

\section{Analysis of Object Detector}
\begin{figure}
    \centering 
    \includegraphics[width=0.22\textwidth]{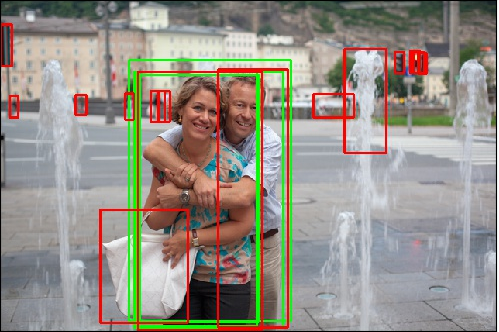}\quad
    \includegraphics[width=0.22\textwidth]{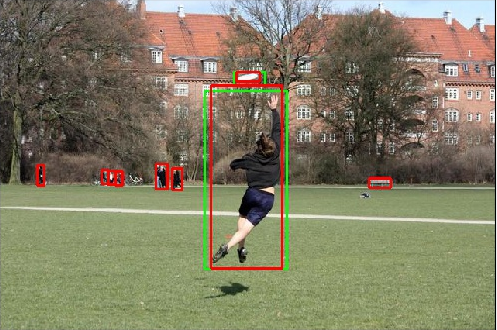}\quad
    \includegraphics[width=0.22\textwidth]{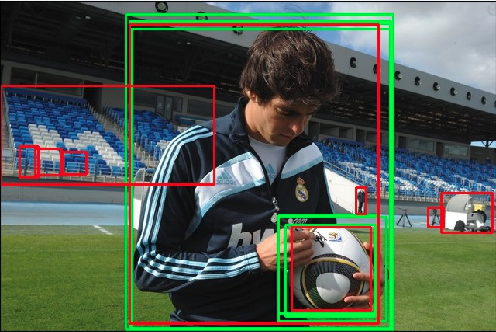}\quad
    \includegraphics[width=0.22\textwidth]{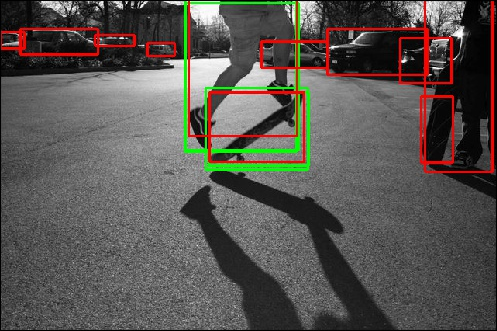}\quad

    \includegraphics[width=0.22\textwidth]{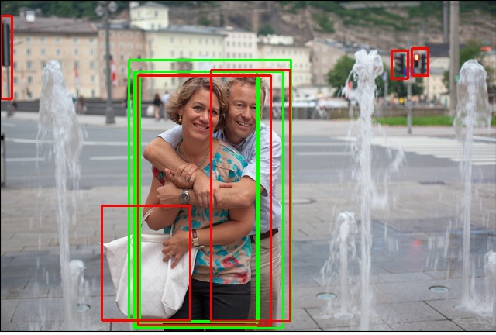}\quad
    \includegraphics[width=0.22\textwidth]{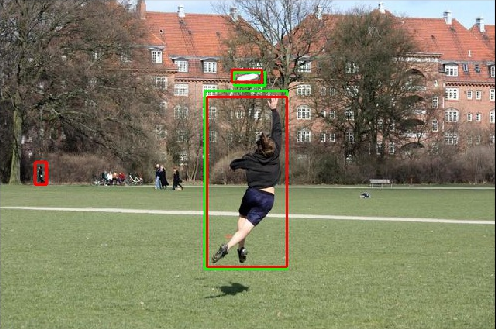}\quad
    \includegraphics[width=0.22\textwidth]{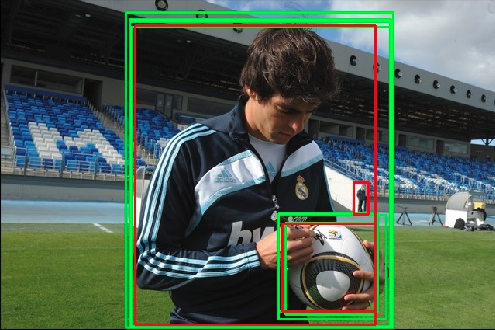}\quad
    \includegraphics[width=0.22\textwidth]{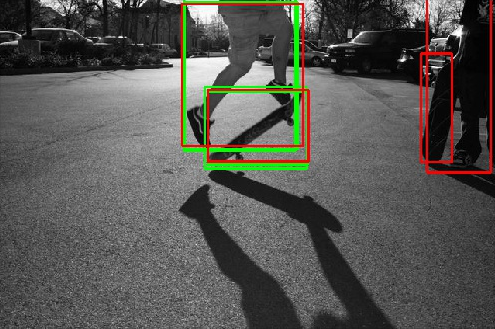}\quad
    \caption{Qualitative Analysis: Comparison of the proposals from (top) a detector trained on COCO
        and (bottom) a detector fine-tuned on objects in HICO-Det. The images shown are from the
        test set of HICO-Det. The green boxes are the ground truth annotations and red boxes
        are the detection outputs. NMS with a threshold of 0.5 was applied on the
    proposals. Detectors fine-tuned on HICO give fewer false positives.}
\label{fig:qual_analysis}
\end{figure}

In this section, we analyze how fine-tuning the object detector on the HICO-Det dataset helps in
improving the HOI detection performance for HICO-Det.  Figure \ref{fig:qual_analysis} shows the
detections on test images of HICO-det for a ResNet-101 Faster R-CNN trained on COCO (top) and HICO
(bottom). For every object class in HICO-det, we select all boxes with confidence score greater than
$0.9$ and apply Non-maximum Supression (NMS) with an IoU threshold of $0.5$ for visualizing the
results. The ground truth boxes are marked in green and the detector outputs are marked in red. 

We observe that annotations of COCO dataset are very tight around the object unlike the HICO-det
dataset. While a detector trained on COCO gives us tight boxes around the object, it can often be
detrimental when evaluated on boxes which are not as tight.  In the HICO-det dataset, most of the
objects in the background have not been labeled because the objects are usually not part of an
interaction, or are out of focus. While detectors trained on COCO densely detect objects in the
image, that can be detrimental for the task of HOI detection on HICO-Det. By densely detecting all
the objects in the image, we increase the number of HOI proposals which in turn results in a large
number of false positives. Using a detector trained for object detection (e.g. COCO) directly for the task of
HOI detection can often lead to inferior results. 

\clearpage

%
%
\bibliographystyle{splncs04}
\bibliography{hoibib}

\end{document}